\documentclass[a4paper,fleqn]{cas-dc}

\usepackage[authoryear,longnamesfirst]{natbib}

\def\tsc#1{\csdef{#1}{\textsc{\lowercase{#1}}\xspace}}
\tsc{WGM}
\tsc{QE}

\begin{document}
\let\WriteBookmarks\relax
\def\floatpagepagefraction{1}
\def\textpagefraction{.001}

\shorttitle{}    

\shortauthors{}  

\title [mode = title]{Bottom-Up Scattering Information Perception Network for SAR target recognition}  

\tnotemark[1] 

\tnotetext[1]{} 

%

\author[1]{Chenxi Zhao}
\author[1]{Daochang Wang}
\author[1]{Siqian Zhang}
\author[1]{Gangyao Kuang}

%
%
%
%


%
%
%
%


\cortext[1]{Corresponding author: Siqian Zhang (zhangsiqian@nudt.edu.cn)}

\fntext[1]{National University of Defense Technology}


\begin{abstract}
Deep learning methods based synthetic aperture radar (SAR) image target recognition tasks have been widely studied currently.
The existing deep methods are insufficient to perceive and mine the scattering information of SAR images, resulting in performance bottlenecks and poor robustness of the algorithms.
To this end, this paper proposes a novel bottom-up scattering information perception network for more interpretable target recognition by constructing the proprietary interpretation network for SAR images.
Firstly, the localized scattering perceptron is proposed to replace the backbone feature extractor based on CNN networks to deeply mine the underlying scattering information of the target.
Then, an unsupervised scattering part feature extraction model is proposed to robustly characterize the target scattering part information and provide fine-grained target representation.
Finally, by aggregating the knowledge of target parts to form the complete target description, the interpretability and discriminative ability of the model is improved.
We perform experiments on the FAST-Vehicle dataset and the SAR-ACD dataset to validate the performance of the proposed method.
\end{abstract}





\maketitle

\section{Introduction}\label{}
\subsection{Background}
Synthetic Aperture Radar (SAR) provides reliable imaging irrespective of weather and time of day, e.g., stable imaging under clouds and fog as well as changes in light.
It plays an important role in the military and civilian fields.
Moreover, with the upgrade of SAR imaging technology, fine-grained target type recognition has progressively emerged as the hot research.

Nowadays, deep learning has become the primary technological tool for fulfilling SAR target recognition tasks.
It liberates humans from the complex manual feature design by completing the feature extraction process autonomously with the help of the annotation information of the data.
Nonetheless, existing deep learning frameworks rely on prior assumptions based on optical imaging principles and data distribution characteristics, which limits their sustained development in the field of SAR target recognition.
The critical impact factor is the inability to extract identifiable and representative features from SAR images, further hindering the model's interpretability.

Existing methods to address such the challenge can be broadly categorized into two groups.
The first group is knowledge supplementation frameworks, which aim to enrich target knowledge representations by modifying the network architecture or incorporating prior target knowledge.
Initially, some researchers designed specialized network architectures tailored to specific characteristics of targets in SAR images.
For example, complete feature extraction of aircraft targets with strong appearance discretization is achieved with the help of deformable convolution and null convolution.
As well, by integrating the multi-scale features of the model to address the problem of large differences in the size of ship targets (e.g., civilian ships and aircraft carriers).
Subsequently, the methods of fusing target auxiliary information gradually emerged as the mainstream in view of the singularity of the deep feature characterization.
Zhang et. al  \cite{9445223} proposed a novel deep learning network with histogram of oriented gradient feature fusion (HOG-ShipCLSNet) for preferable SAR ship classification. 
Furthermore, the target shadow information which is widely used in SAR target recognition tasks.
As more and more researchers realize the importance of electromagnetic scattering features, methods for fusing the features in the network have received widespread attention and have been validated in aircraft, ship, and vehicle target recognition tasks.
In summary, this group of methods works to integrate diverse knowledge about the target to strengthen the target representation.
However, the following problems remain:
i) Different recognition tasks rely on proprietary networks and have poor model stability and generalization.
ii) The model relies on accurate a priori feature knowledge and effective feature fusion strategies which have poor model reliability.

The second category is knowledge enhancement architectures, specifically designed to construct target representations with strong discriminative capabilities.
By constructing intermediate features with strong representation capabilities, the discriminability and interpretability of target representations are enhanced.
Discriminative knowledge is usually present in local parts of the target, such as gun barrels and tracks. Enhancing attention to local details will improve the discriminability and robustness of features.
Specifically, Feng et. al \cite{10068265,9896887} and Huang et. al \cite{HUANG2020179} clustered scattering center features to generate target scattering parts information, and then aggregated parts knowledge by LSTM, attention mechanisms, etc., to generate complete target representations.
Furthermore, some other scholars have improved the separability or tightness of local features by capturing the current critical image region through automatic segmentation, which allows the model to capture more critical parts of the image and extract more effective features.
The exploration of component-level knowledge has always been a research focus, as it not only enhances the discriminability of target descriptions but also improves their interpretability.
Existing methods still rely on prior knowledge for part annotation, which limits the stability and generalization of the algorithms.


\subsection{Motivation}
In general, the main problems with existing methods are as follows:
\begin{itemize}
	\item[1)] There is still no escape from the inherent incompatibility that exists between existing deep network structures and SAR images.
	\item[2)] The strong dependence on a priori knowledge leads to poor model generalization and stability.
\end{itemize}

These questions prompted the following reflections:
\begin{itemize}
	\item[1)] On the scattering information aware proprietary network. The imaging mechanisms of electro-optical (EO) and SAR images are fundamentally distinct.
	The SAR image is the quantitative indication of the target's radar cross-section (RCS) echo energy and thus exhibits as the gray image.
	Moreover, the texture on the SAR image originates from the backscattering result of the local structure on the target, which is the essential distinction with the EO image.
	Thus, compared to EO images, SAR images provide less image information and poor visual interpretability.
	Existing excellent deep learning frameworks, such as CNN, transformer, etc., which are inherently developed on the principle of optical image imaging, are hardly to capture the essential information about the SAR target effectively, which leads to poor model interpretability and generalization ability.
	To break out of the dilemma of SAR image target recognition, the proprietary network for SAR image feature extraction that genuinely reflects the intrinsic properties of the target must be developed separately from the inherent properties of SAR images.
	\item[2)] On the discriminative scattering parts discovery. 
	Target part information contains more fine-grained information, which enhances the discrimination of target description on the one hand, and promotes the interpretability of the features and the reliability of the model on the other hand.
	The definition and acquisition of parts are critical issues, and the common approach is to leverage prior electromagnetic scattering features to generate the scattering part features.
	However, it is evident that such a straightforward approach to part generation is dependent on high-quality a priori features.
	Therefore, the complexity and the generalizability of these methods are up for consideration.
	In light of this, the development of a smarter, visually interpretable SAR part discovery method that does not involve the incorporation of extra information (auxiliary features or labeling information) is imminent.
\end{itemize}


The bottom-up scattering information perception (BU-SIP) network is proposed to solve the critical problem of SAR image target recognition from the above issues.

\subsection{Contribution}
Specifically, the BU-SIP framework hierarchically extracts knowledge about the target's scattering properties—from the perception of low-level scattering information to the recognition of higher-level scattering parts—thereby enhancing interpretability during classification.
With the powerful local analysis ability of wavelet transform on images and the rich wavelet basis, the multi-scale representation and accurate analysis of local scattering phenomena on SAR images will be realized.
Thus, we propose the first proprietary interpretation network for SAR image properties.
Subsequently, the unsupervised scattering part representation module is proposed to acquire the target part representation without any additional annotation information.
Finally, the scattering part knowledge is aggregated to obtain the complete and more discriminative target representation.
In conclusion, the proposed method not only improves the discriminative ability of the model, but also has strong interpretability.
The contributions of our work are summarized below:
\begin{itemize}
	\item[1)] A novel model of bottom-up scattering information perception network in image domain is proposed in this paper.
	To hierarchically implement from bottom scattering information perception to high-level scattering part discovery, extract discriminative target features and enhance model interpretability.
	\item[2)] A proprietary image interpretation network structure for SAR is proposed for the first time.
	The wavelet transform is the theoretical basis and the wavelet scattering network is the technical basis to realize the perception of the bottom scattering information of SAR images.
	\item[3)] The scattering part representation network and the part knowledge aggregation network are proposed.
	Firstly, the target scattering part discovery is realized without any labeling information.
	Upon this basis, the parts knowledge are aggregated to generate the complete target description.
	Ultimately, the output consists of information about the target category and the contribution of the various parts to achieve an interpretable classification process.
\end{itemize}

The remainder of the paper is organized as follows. We first review the related works in Section II; Section III describes the proposed method in detail; Section IV reports extensive experimental results and analysis; and finally the paper is summarized in Section V.

\section{Related Work}

\subsection{Scattered information awareness}

Wavelet transform (WT), with multi-scale and localization properties, is the major mathematical tool for analyzing images, especially in dealing with non-smooth signals and complex textured scenes with irreplaceable advantages.
Given these properties, the WT is widely used in SAR image interpretation tasks.
Grandi et al. \cite {4695998, 4358832} presented the theoretical approach to derive texture measurements from Synthetic Aperture Radar (SAR) data via wavelet frames.
Moreover, experiments are reported aiming at characterizing the sensitivity of wavelet texture measurements to target structural properties and SAR configurations from a purely observational point of view.
Tello et. al \cite{1420305} exploited the differences in statistics between ships and the surrounding sea to provide more reliable ship detection by interpreting the information through wavelet coefficients.
Duan et. al \cite{DUAN2017255} designed the wavelet pooling layer to replace the conventional pooling in CNN, which can suppress the noise and is better at keeping the structures of the learned features. 
Therefore, the WT is a very suitable mathematical tool for analyzing the local scattering phenomenon in SAR images.

Given that WT has excellent image analysis capabilities, it is naturally to integrate it into a deep network framework.
Wavelet Scattering Network (WSN) is the wavelet transform-based deep feature extraction architecture proposed by Mallat et al. \cite{6522407} firstly, which intends to extract features with deformation invariance through multiscale analysis and nonlinear transforms.
With the continuous research \cite{6619007, 7298904,7822922}, WSNs based on different wavelet basis and various model structures have been proposed.
Dong et. al \cite{7894231}  developed a new representation model via the steerable wavelet frames. The proposed representation model allows for invariance towards translations and rotations, and hence provides great potential for target recognition.
The above networks are a class of designed Convolutional Neural Networks (CNNs) with fixed weights. We argue they can serve as generic representations for modelling images. 
Subsequently, some scholars proposed hybrid models (WSN and CNN)\cite{8413168} and parameter learnable \cite{9878805} WSN frameworks for better image representation.
In such research context, it is worthwhile and feasible to design network structure suitable for SAR image analysis.

\subsection{Target part discovery}
Unsupervised part discovery intends to extract part information about the target without extra part annotation.
It is also the popular research direction in the field of computer vision, such as object classification and person re-identification.
Early works \cite{8237610, 8953229,9157763,jin2022tusk} focused on unsupervised landmark inspection aimed at obtaining part location information.
Such tasks are simple and direct, inherently treating the landmark of the target as the simplified form of part-level labeling.
However, as the study progressed, the researchers found that target landmark detection is insufficient to characterize the various parts of the target \cite{9157722}.

Subsequently, some approaches attempt to extract part knowledge through auxiliary constraint information.
For example, Collins et al. \cite{Collins2018} utilized matrix decomposition to highlight regions with similar semantics in the feature map, roughly delineating information about different parts.
Liu et al.\cite{9578754} introduce shape and appearance constraints on the target and display extracted part representations, but the learned representations do not adequately describe the corresponding parts.
Overall, these methods fail to learn high-quality part representations.

On the other hand, vision transformer (VIT) \cite{vaswani2017attention} can learn instance-level attention without labeling and extract high-quality instance-level representations to facilitate downstream tasks.
The biggest advantage of VIT-based part discovery methods \cite{10638824,10106785} is the availability of global contextual information about the target, which is the potential method for aggregating similar features to adaptively generate high-quality part representations.
Li et. al \cite{Li_2021_CVPR} proposed a novel end-to-end Part-Aware Transformer (PAT) for occluded person Re-ID through diverse part discovery via a transformer encoder-decoder architecture.
Ni et. al \cite{Ni_2023_ICCV} proposed a pure Transformer model for domain generalization person re-identification by designed a proxy task,  named Cross-ID Similarity Learning (CSL), to mine local visual information shared by different IDs. 
We propose a novel scattering part discovery module and constrain the learning process by multiple loss functions to obtain target scattering parts with orthogonality and completeness.
The model training process employs only the target categories as supervisory information, which not only enables the generated part features to serve the final recognition task, but also enhances the semantic interpretability of the part representation.

\section{Methods}

\begin{figure*}[!t]
	\centering \includegraphics[width=6.8in]{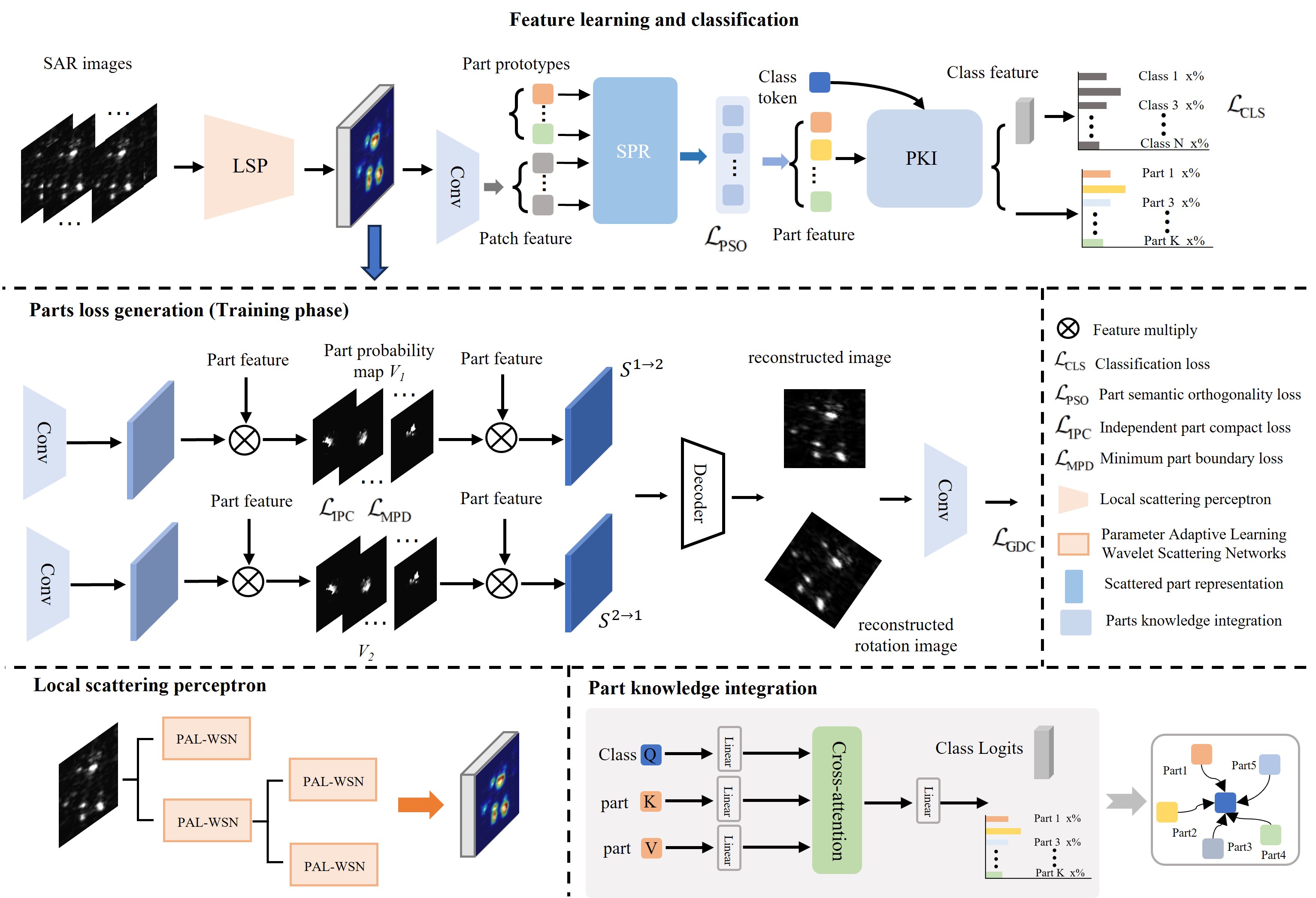}%
	\caption{BU-SIP network structure}
	\label{fig_oa}
\end{figure*}

In this paper, a novel bottom-up scattering information perception network for SAR target recognition is proposed, as shown in Fig. \ref{fig_oa}, consisting of three crucial modules: the local scattering perceptron, the scattering part representation module, and the part knowledge aggregation module.
In this paradigm, high-performance recognition results are obtained and they are extremely interpretable from the feature level to the decision level.

\subsection{Overall Framework}
The model takes the SAR image as the only input.
Firstly, the local scatter perceptron is employed to extract the knowledge of the bottom scattering properties of SAR image, which is the basis of the subsequent tasks.
Subsequently, the target bottom scattering information is further aggregated to form the high-level target scattering parts representation with stronger feature discrimination and interpretability.
Finally, adaptive aggregation of target parts knowledge to generate more discriminative holistic representations of the target, and the contribution of various parts to the final result can be captured.
On the one hand, it motivates the network to focus more locally on the target scattering parts information to provide richer description about the target to the network.
On the other hand, it provides interpretability to the deep learning model, especially about the decisional process or the predicted outputs, which allows to understand the logic behind it, and thus ensures the trust and transparency to the network.

Specifically, during the training process, we employ the data augmentation technique of geometric transformations to enhance the generalization ability of the model.
This is attributed to the fact that geometric transformations have been proven to be the simple and powerful supervised signals in unsupervised representation learning.

Notably, we propose a model for sensing multi-level scattering information of the target from the image domain for the first time.
The bottom-up extraction of target scattering representations leads to high-performance recognition results and strong interpretability.


\subsection{Local scattering perceptron}
%

WT has the strong ability about time-frequency localization and can precisely locate the mutation point or edge information of the signal.
Thus, it is the powerful tool to analyze the local scattering phenomenon from the SAR image.
In this section, the local scattering perceptron is described in the following three aspects:
i) Firstly, the local scattering phenomenon of SAR images and its image domain modeling form are analyzed.
ii) Based on the target local scattering model, the wavelet basis function is selected and its performance is analyzed.
iii) Finally, the basic wavelet scattering network structure used in this paper is presented.

\subsubsection{Analysis of localized scattering phenomena}
\begin{figure}[!t]
	\centering \includegraphics[width=3.3in]{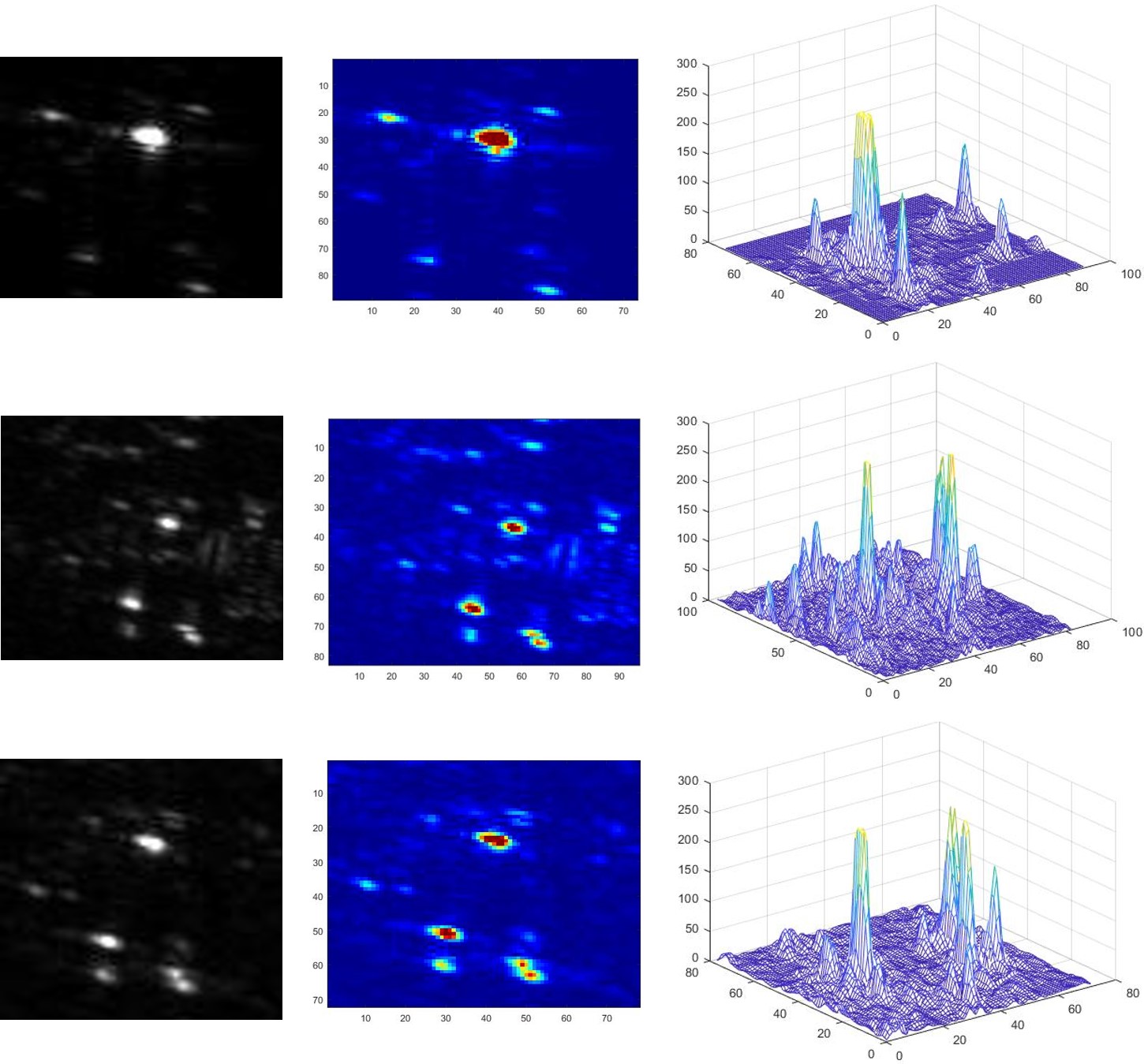}%
	\caption{SAR image peak value visualization results}
	\label{fig30}
\end{figure}

Fig. \ref{fig30} shows the SAR image, the color mapping of the image, and the three-dimensional presentation of the data.
Obviously, from the three-dimensional visualization of the SAR image, it is found that the contours of the local scattering region approximate circles or ellipses.
Therefore, it can be described by two-dimensional (2D) Gaussian distribution model.

The most important prerequisite for analyzing the image by wavelet transform is the selection of the suitable wavelet basis function.
The selection of the wavelet basis needs to follow the following criteria:
i) Signal properties. The wavelet basis is selected according to the properties of the signal. 
For example, if the signal contains fast mutation or intermittent features, select Hart wavelet or Daubechies wavelet.
ii) Application Requirements. For example, the wavelet basis is required to have superior symmetry and tight branching in image processing tasks, whereas signal compression focuses more on orthogonality and computational efficiency.
iii) Computational efficiency. Some wavelet basis (e.g., Hart wavelet) are computationally simple, but may not accurately describe complex signals.
For applications that demand computationally efficient wavelet bases, Hart wavelets or low-order Daubechies wavelets can be excellent choices.

Specifically, criterion 1) is the most important selection criterion.
Combined with the 2D Gaussian model modeling manner of local scattering, 2D Morlet wavelets are selected for SAR image analysis in this paper.

\subsubsection{Morlet wavelet-based SAR image analysis}
The 2D Morlet wavelet formula is shown below,
\begin{equation}\label{eq1}
	\psi(x, y)=e^{-(\frac{x^2}{2 \sigma_x^2}+\frac{y^2}{2 \sigma_y^2})} \cdot e^{i k_0(x \cos \theta+y \sin \theta)}
\end{equation}
where $x, y$ indicates spatial coordinates.
$\theta$ denotes the azimuth, which determines the orientation of the wavelet in space.
$k_0$ is the center frequency of the wavelet.
$\sigma_x, \sigma_y$ represent the spread of the Gaussian envelope in the $x$ and $y$ directions, respectively, which determines the spatial localization of the wavelet (the larger $\sigma$, the wider the coverage).
Thus, by changing the $\sigma$ and $\theta$ parameters in the morlet wavelet, it is possible to generate wavelet basis with various scales and orientations.

\begin{figure}[!t]
	\centering \includegraphics[width=3.3in]{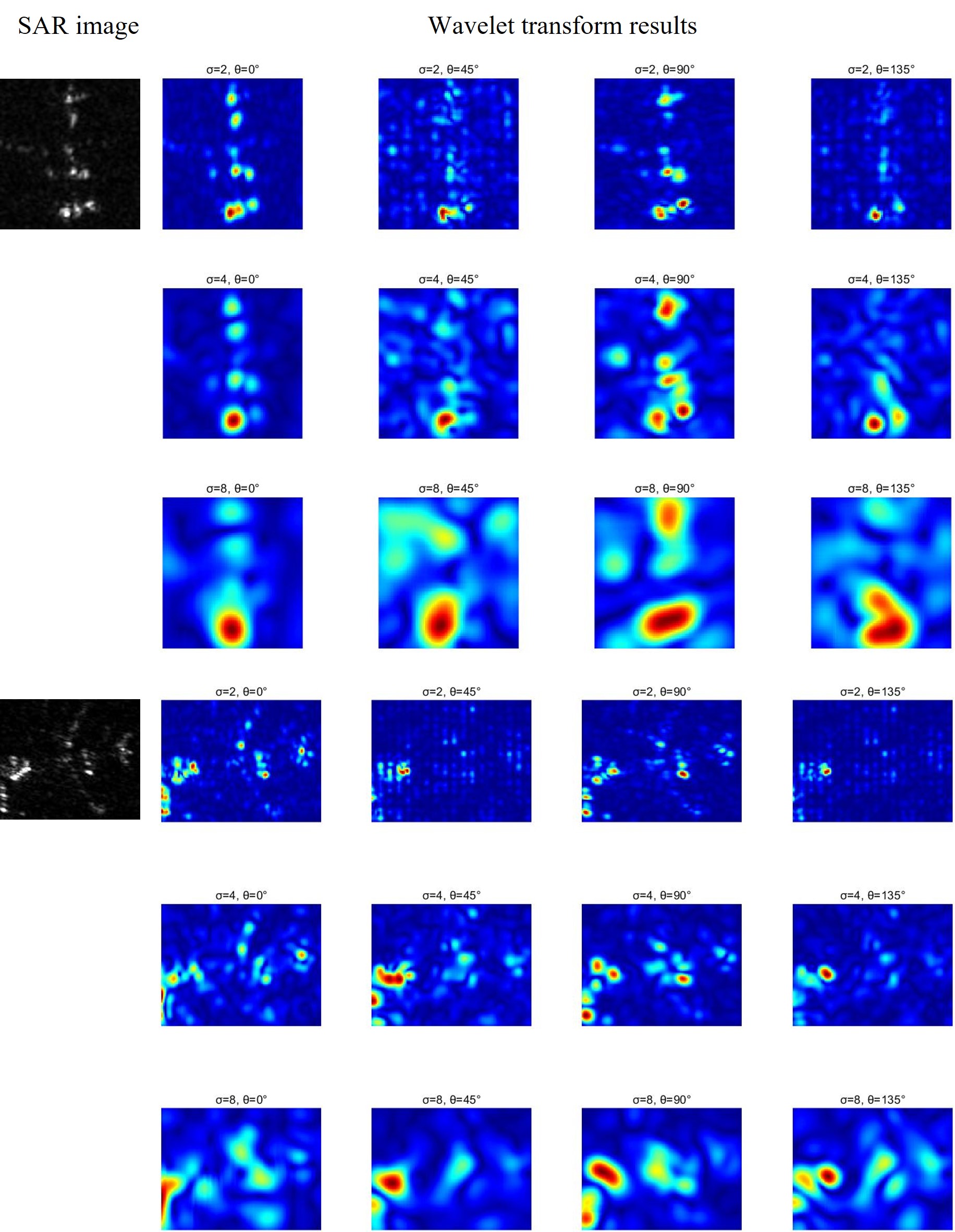}%
	\caption{Morlet wavelet processing SAR image visualization results}
	\label{fig31}
\end{figure}
Here, we illustrate the analysis of SAR images employing morlet wavelets, as shown in Fig. \ref{fig31}.
Obviously, with different combinations of $\sigma$ and $\theta$ parameters, various image detail information will be preserved.

\subsubsection{Network Structure}
WSN is the feature extraction method that combines the ideas of WT and deep learning.
Conventional wavelet scattering networks rely on fixed wavelet bases and non-tunable parameters, which brings good theoretical stability and robustness but lacks the ability to adapt to specific data or tasks.
The tunable parameters permit the scattering network to maintain both theoretical interpretability and flexibility, and thus play the greater role in data-driven feature learning.
For this reason, the parametric wavelet bases are utilized in this paper, replacing the traditional fixed wavelet bases with parametric filters whose parameters such as shape, orientation, and bandwidth can be optimized through learning.

Equation (\ref{eq1}) is a simple Morlet wavelet function form, as the local scattering information has certain directionality and multi-scale nature, the specific representation is employed as:
\begin{equation}
	\psi_{\sigma, \theta, \xi, \gamma}(u)=e^{-\left\|D_\gamma R_\theta(u)\right\|^2 /\left(2 \sigma^2\right)}\left(e^{i \xi u^{\prime}}-\beta\right)
\end{equation}
where $\beta$ is a normalization constant to ensure that the wavelet integrates to 0 in the spatial domain.
$u^{\prime}=u_1cos\theta+u_2sin\theta$, $R_{\theta}$ is the rotation matrix for angle $\theta$, $D_\gamma=\left(\begin{array}{ll}1 & 0 \\ 0 & \gamma\end{array}\right)$.
These four parameters can be adjusted as shown in Table \ref{tab31}.
\begin{table}[h!]
	\caption{Learnable Parameter Name}
	\label{tab31}
	\centering
	\setlength{\tabcolsep}{1mm}{
	\begin{tabular}{c c c c c}
		\hline
		\textbf{Parameter} & $\sigma$  & $\theta$ & $\xi$   & $\gamma$      \\ \hline
		\textbf{Name} & Gauss size  & Orientation    & Frequency    & Aspect ratio    \\ \hline
	\end{tabular}
}
\end{table}
Specifically, the Morlet wavelet that best fits the distribution of SAR image data can be learned by optimizing the parameters in Table \ref{tab31}.
Furthermore, we adopt the non-downsampling pattern of wavelet scattering network, which effectively avoids the artifacts or distortions caused by downsampling in the traditional wavelet transform, and better captures the local detail information.

\subsection{Scattering part representation}
After obtaining the essential scattered knowledge of SAR images, the unsupervised scattered part representation module is proposed in this paper.
In the absence of additional supervisory information, details regarding the target's scattered parts can still be extracted.

The module is based on VIT and replaces the token with $K+1$ part prototypes and $N$ image patches.
The final output contains $K$ part representations, and $1$ background representation.
\begin{equation}
	\{p_0^i, p_1^i,..., p_K^i\}=SPT(\{T_0, T_1,...,T_K,p_0^i, p_1^i,..., p_{N-1}^i\})
\end{equation}

\subsection{Parts Knowledge Aggregation}
With the scattered parts representations of the target acquired, the parts knowledge aggregation module is proposed to implement the aggregation of the target parts knowledge to form the complete target description.

The module is based on VIT and is designed with the help of the cross-attention mechanism, as shown in Fig. \ref{fig_oa}.
Cross-attention is an important variant of the attention mechanism that is mainly used to handle the associative relationship between two sequences.
In this module, $K$ part features, and one category token $C_0$ are input, and the final output are the category representations and the scoring about each part.
\begin{equation}
	f_{cls}, P^{att}=\operatorname{PKI}\left(\left\{C_0, p_1, p_2, \ldots, p_K\right\}\right)
\end{equation}
Then, the category features $f_{cls}$ are fed into a fully connected classifier that predicts the target category $\hat{y}$.
Ultimately, the optimized objective function can be expressed as:
\begin{equation}
	\mathcal{L}_{\mathrm{cls}}=CE(\hat{y},y)
\end{equation}
where CE denotes the cross-entropy loss function and $y$ denotes the true category of the target.


Obviously, the method captures the essential scattering information of the SAR image target in the bottom-up manner and realizes the more interpretable feature extraction process.

\subsection{Part Loss function}
The multiple loss functions are proposed for constraining the model training process to obtain the better description of the scattering parts.

\subsubsection{Independent part compact loss}
The compactness scattering part description can be obtained by aligning the part pixels with the semantic information to make the semantic part representation with high similarity to the pixel representation of the corresponding part region and low similarity to the pixel representation of the other regions.
Furthermore, the foreground pixels with the same semantics on the probabilistic map $V$ are prompted to form centrally connected portions, and the scattering parts are found to be locally focused by the modules and form portions of specific attention.
The independent parts compact (IPC) loss is expressed as follows:
\begin{equation}
	\mathcal{L}_{\mathrm{IPC}}=\frac{\sum_{k=1}^K \sigma_x^2\left({V}^k\right)+\sigma_y^2\left({V}^k\right)}{K}
\end{equation}
where $\sigma_x^2$ and $\sigma_y^2$ denote the variance of the $k_{th}$ part region vertically and horizontally, respectively.

\subsubsection{Minimum part boundary loss}
The minimum part boundary (MPB) loss limits the minimum area size of the foreground part and background on $V$.
If left unconstrained during training, the concentration constraints will be prioritized by describing the entire image using a background representation.
Thus, the training result becomes a mundane solution.
The minimum part boundary loss is expressed as follows:
\begin{equation}
	\mathcal{L}_{\text {MPD}}=\sum_{k=1}^{K+1} \frac{1}{1+z_k / \alpha}
\end{equation}
where $\alpha$ is a hyperparameter.
The $z_k=\sum_{i=1,j=1}^{W,H} V_{i,j}^{k}+\epsilon$ is a prediction for the $k_{th}$ component region and $\epsilon$ is a small constant in case $z_k$ is zero.
When the area $z_k$ of the prediction region is less than the threshold, $\mathcal{L}_{\text {MPD}}$ will increase rapidly.
Therefore, when the predicted area size will be larger than this threshold, this will prevent the model from entering a mundane solution.
This threshold is determined by $\alpha$ and therefore $\alpha$ is used as a priori knowledge to determine the desired minimum size of each predicted region.

\subsubsection{Global description consistency loss}
The global description consistency (GDC) loss constrains the part acquisition process potentially by penalizing the difference between the input image and the reconstructed image.
Specifically, the variability between features is measured instead of directly measuring the differences between images.
In view of the superiority of VGGNet network in the field of SAR feature extraction, VGGNet is adopted as the backbone network $\Phi$ for feature extraction of the image, and then the reconstruction loss $\mathcal{L}_{\text {GDC}}$ is calculated:
\begin{align}
	\mathcal{L}_{\text {GDC }}=& cosine(\Phi\left(\boldsymbol{I}^{(1)}\right),\Phi\left(\boldsymbol{I}^{\prime(1)}\right))\\
	& + cosine(\Phi\left(\boldsymbol{I}^{(2)}\right),\Phi\left(\boldsymbol{I}^{\prime(2)}\right))
\end{align}
Since pairwise feature alignment is used in the specific execution of the algorithm, the information for each part region in $S^{1\rightarrow 2}$ and $S^{2\rightarrow 1}$ is mainly derived from the part representations of their responses.
Thus, the constraint encourages that the information in each part region of the original input data is shown to be included in the corresponding part representation.
Moreover, the reconstructed features should also be consistent with the part representations extracted from the original input, which ensures better geometric transformation stability of the learned part representations.
The cosine similarity metric can effectively perceive the SAR image target information, therefore, the cosine metric $cosine$ is employed here.

\subsubsection{Part Semantic Orthogonality loss}
Finally, the part semantic orthogonality (PSO) loss is intended to ensure that distinct part representations in the feature space are orthogonal, and to induce the same part representation close in the potential feature space.
Given this, the ArcFace loss based on the cosine metric is employed to guarantee the semantic consistency of the parts.
Specifically, $K$ learnable vectors of dimension $C$ are defined for $K$ foreground parts, which can be written as $W\in \mathbf{R}^{K\times C}$.
$W$ is the shared learnable weight.
The angle between the $t$th vector of $W$ and the $k$th representation of $G$ is defined as $\theta(t,k)$.
The part independence loss $\mathcal{L}_{PSO}$ is then computed as:
\begin{equation}
	\mathcal{L}_{\mathrm{PSO}}=-\frac{1}{K} \sum_{k=1}^K \log \frac{e^{s\left(\cos \left(\theta_{(k, k)}+m\right)\right)}}{e^{s\left(\cos \left(\theta_{(k, k)}+m\right)\right)}+\sum_{t=1, t \neq k}^K e^{s \cos \theta_{(t, k)}}}
\end{equation}
where $m$ and $s$ are two hyperparameters that penalize corner edges and scaling radius.
On the one hand, the part semantic orthogonality loss encourages each partial representation of all training samples to approximate its corresponding learnable vector by minimizing $\theta_{(k,k)}$ over the cosine distance.
As a result, a more consistent semantics is obtained for representations of the same part extracted from different images.
On the other hand, the PSO loss maximizes $\theta_{(k,k),(t\neq k)}$ to encourage the representations of the different parts to be orthogonal.
This makes it easier to distinguish between different part representations in the potential space.

\subsubsection{Total loss}
The proposed method consists of three important modules: the local scatter perceptron, the scattered part representation module, and the part knowledge integration module.
Specifically, superior scattered part representations are generated through the joint constraints of independent part compact loss, minimum part boundary loss, global description consistency loss and part semantic orthogonality loss.
Finally, all parts knowledge are integrated to achieve target identification.
\begin{equation}
	\mathcal{L}=\mathcal{L}_{\text {GDC}}+\lambda_{\mathrm{PSO}} \mathcal{L}_{\mathrm{PSO}}+\lambda_{\text {IPC}} \mathcal{L}_{\text {IPC}}+\lambda_{\text {MPD}} \mathcal{L}_{\text {MPD}}+\mathcal{L}_{\mathrm{cls}}
\end{equation}
where $\lambda_{\mathrm{PSO }}$, $\lambda_{\text {MPD}}$ and $\lambda_{\text {IPC}}$ are the corresponding weights of $\mathcal{L}_{\text {GDC }}$, $\mathcal{L}_{\mathrm{PSO}} $ and $\mathcal{L}_{\text {IPC}}$ correspondingly.

\section{Experiments and analysis}
Our experiments are performed on the personal computer (PC) equipped with Intel i9-11900K CPU, NVIDIA
RTX3080Ti GPU and 32G RAM. The experiments are implemented in the Python environment, with the open-source machine learning library Pytorch. Moreover, CUDA 11.3 is used to accelerate the computation process by GPU.

\subsection{SAR Dataset}
To fully validate the effectiveness of the method, this paper performs experimental validation on two existing public datasets of SAR images, Full aspect stationary targets-Vehicle (FAST-Vehicle) and SAR aircraft category (SAR-ACD).

\subsubsection{Fast-Vehicle}
\begin{figure*}[!t]
	\centering
	\includegraphics[width=5.5in]{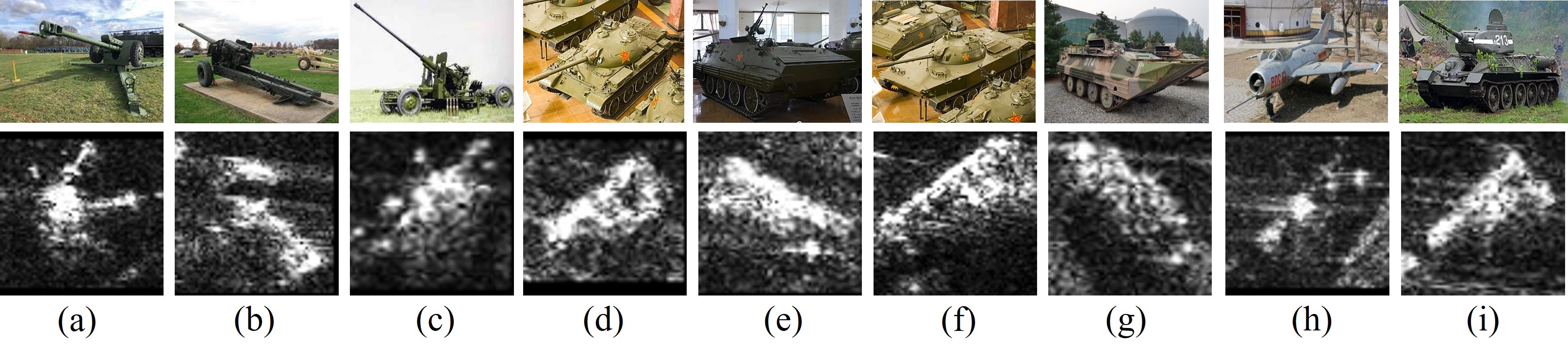}
	\caption{Optical images and corresponding SAR images of nine types targets in the FAST-Vehicle dataset. (a) 54TH. (b) 591TC. (c) 59AG. (d) 62LT. (e) 63APTV. (f) 63AT. (g) 63CAAV. (h) J6. (i) T3485.}
	\label{fE1}
\end{figure*}

Full aspect stationary targets-Vehicle (FAST-Vehicle) is the vehicle SAR data in X-band spot mode recorded by MiniSAR system researched and developed by Nanjing University of Aeronautics and Astronautics (NUAA), with the resolution of 0.1 meter and the azimuth angle from 0 to 360.
The SAR images are captured in March and July 2022 include nine categories, namely 62LT, 63APTV, 63CAAV, 63AT, T3485, 591TC, 54TH, 59AG, and J6.
The optical image of the above targets and their corresponding SAR images, are shown in Fig. \ref{fE1}.

Compared to the classical SAR ATR dataset, Moving and Stationary Target Acquisition and Recognition (MSTAR), FAST-Vehicle contains SAR image data under richer conditions, including various imaging parameters, and various collection times, as shown in Table \ref{tab1}.
To more fully verify the validity of the proposed method, we set up three main groups of experimental conditions, which are:
\begin{itemize}
	\item[1)] Scenario 1: same time, different depressions. July $26^{\circ}$ depression angle data is used as the training set, and $31^{\circ}$, $37^{\circ}$, and $45^{\circ}$ are used as the test set, respectively.
	\item[2)] Scenario 2: different time, same depression. We treat July $31^{\circ}$ and $45^{\circ}$ as the training set and correspondingly March $31^{\circ}$ and $45^{\circ}$ as the test set, respectively.
\end{itemize}

\subsubsection{SAR-ACD}
\begin{figure*}[!t]
	\centering
	\includegraphics[width=5.3in]{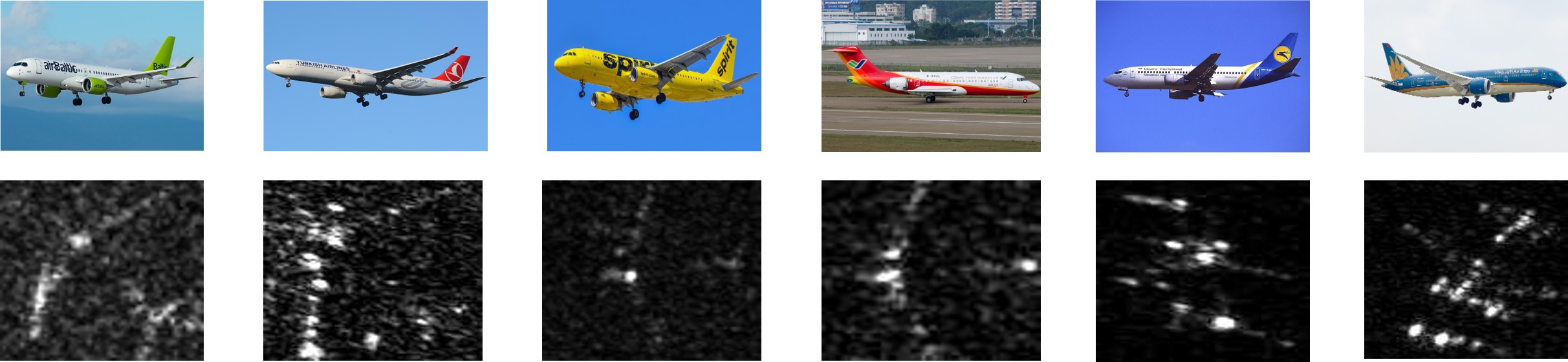}
	\caption{Optical images and corresponding SAR images of nine types targets in the SAR-ACD dataset. (a) A220. (b) A330. (c) A320/321. (d) ARJ21. (e) Boeing 737. (f) Boeing 787.}
	\label{fE1-1}
\end{figure*}

SAR aircraft category (SAR-ACD) dataset is the dataset acquired by the Gaofen-3 satellite in the C-band.
The SAR-ACD dataset includes 6 civil aircraft categories and 14 other aircraft categories from the Shanghai Hongqiao Airport area and other airports.
The civil aircraft category consists of the A220, A320/321, A330, ARJ21, Boeing 737 and Boeing 787.
The optical image of the above targets and their corresponding SAR images, are shown in Fig. \ref{fE1-1}.
This paper focuses on the six known classes civil aircraft recognition.
Table \ref{tab2} illustrates the training and test set allocations for the experiments in this paper.

\begin{table*}[!t]
	\renewcommand\arraystretch{1.2}
	\setlength\tabcolsep{1mm}
	\caption{Details of the FAST-Vehicle dataset\label{tab1}}
	\centering
	\begin{tabular}{c c c c c c c | c c c c}
		\hline
		 \multirow{4}{*}{Class}     &\multicolumn{6}{c|}{Scene 1}       &\multicolumn{4}{c}{Scene 2} \\
		 \cmidrule(r){2-7} \cmidrule(r){8-11}
		                          &\multicolumn{2}{c}{EXP1}  &\multicolumn{2}{c}{EXP2}  &\multicolumn{2}{c|}{EXP3}  &\multicolumn{2}{c}{EXP4}    &\multicolumn{2}{c}{EXP5}  \\
		                          \cmidrule(r){2-3} \cmidrule(r){4-5}  \cmidrule(r){6-7}  \cmidrule(r){8-9} \cmidrule(r){10-11} 
		                          &Train &Test    &Train  &Test   &Train  &Test   &Train &Test    &Train  &Test   \\
		                          &July $26^{\circ}$ &Mar $31^{\circ}$   &July $26^{\circ}$ &Mar $37^{\circ}$
		                          &July $26^{\circ}$ &Mar $45^{\circ}$   &July $31^{\circ}$ &July $31^{\circ}$
		                          &July $45^{\circ}$ &July $45^{\circ}$  \\
		                          \hline
		54TH     &231 &286   &231 &295   &231 &282      &286 &172   &282 &128    \\
		59-1TC   &231 &286   &231 &297   &231 &291      &286 &307   &291 &270    \\
		59AG     &228 &283   &228 &293   &289 &291      &283 &821   &289 &727    \\
		62LT     &230 &286   &230 &296   &230 &291      &286 &150   &291 &129    \\
		63APTV   &230 &286   &230 &296   &230 &289      &286 &158   &289 &122    \\
		63AT     &228 &286   &228 &296   &228 &292      &286 &185   &292 &161    \\
		63CAAV   &231 &286   &231 &296   &231 &291      &286 &157   &291 &104    \\
		J6       &230 &286   &231 &293   &231 &292      &286 &174   &292 &115    \\
		T34-85   &222 &278   &222 &295   &222 &289      &278 &718   &289 &551    \\
		\hline
	\end{tabular}
\end{table*}

\begin{table}[!t]
	\renewcommand\arraystretch{1.6}
	\caption{Details of the SAR-ACD dataset} \label{tab2}
	\centering
	\setlength{\tabcolsep}{5mm}
	{
		\begin{tabular}{c  c  c }
			\hline
			Class                & Train             & Test   \\
			\hline
			
			A220                & 200       & 264     \\
			A320/321            & 200       & 310     \\
			A330                & 200       & 312     \\
			ARJ21               & 200       & 314     \\
			Boeing 737          & 200       & 328     \\
			Boeing 787          & 200       & 304     \\
			
			\hline
		\end{tabular}
	}
\end{table}

%
%
%
%

\subsection{Local scattered perceptron analysis}
\begin{figure*}[!t]
	\centering \includegraphics[width=6.3in]{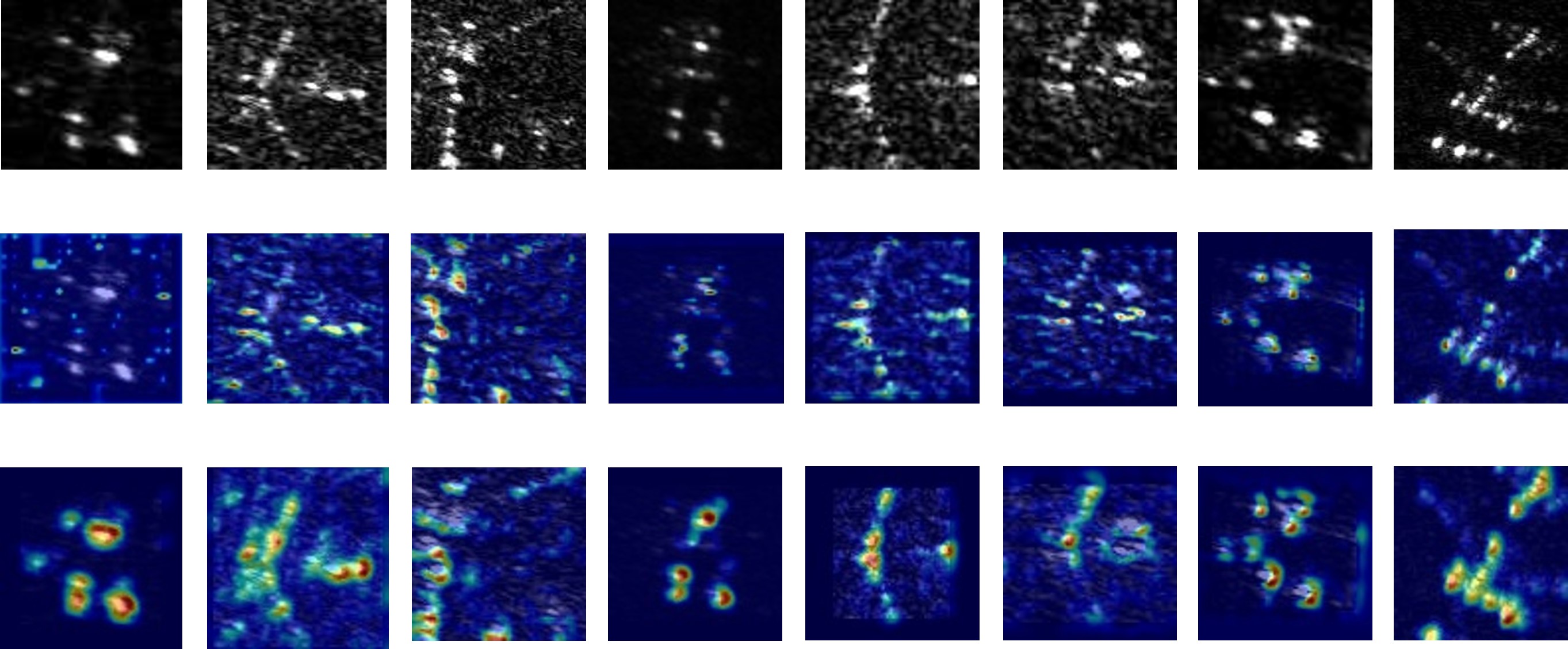}%
	\caption{Aircraft target feature visualization results.}
	\label{fig53}
\end{figure*}

Here, we visualize and analyze the ability of the local scattered perceptron to capture the scattered information from the target, as shown in Fig. \ref{fig53}.
Given the strong appearance discretization of aircraft targets in SAR images, the superiority of the method is better demonstrated on the aircraft target dataset.
Specifically, we will visualize the features extracted by the CNN-based network versus those extracted by the wavelet scattering-based network to visualize the network's ability on capturing scattered information in SAR images.

Obviously, compared with the CNN network-based method, the local scattered perceptron model can easily perceive the target local scattered information, and the large amount of noise information in the background are effectively suppressed.
This is exactly the network structure required for SAR image interpretation.

The background noise and target local scattering representations obey different distributions, and by using the wavelet scattering network, the target local scattered features can be strengthened on the one hand, and the influence of background noise can be effectively suppressed on the other hand.

\subsection{Ablation experiments}
In this paper, several modules with loss functions are proposed to achieve the more complete and independent target scattered part representations.
Thus, it is essential to explore the effectiveness of these modules and losses.
It is worth mentioning that the ablation experiments are all completed in the FAST-Vehicle EXP1 condition.

\subsubsection{Module validity analysis}
The baseline method is AconvNet, the SAR ATR-specific network, which achieves the recognition accuracy of 59.19\%.
Subsequently, the effectiveness of the scattered part representation module and the part knowledge integration module in the recognition task is further analyzed.
Obviously, when the target part description has been obtained, it is possible to achieve a significant performance improvement up to 13.54\%.
This result demonstrates that part information will provide more fine-grained target knowledge and enhance feature discriminability.
Subsequently, the part knowledge integration module is introduced into the network, which utilizes the cross-attention mechanism to quantify the role of each part in the final recognition task, thus enhancing the expressiveness of the features.
Ultimately, the recognition accuracy reached 75.69\%.
Evidently, fine-grained part knowledge for SAR images facilitates the ability of the model to improve the comprehension of the image and thus obtain more discriminative target representations.

Then, we replace the feature extraction backbone network with the local scattered perceptron to deeply analyze the effect of target scattered knowledge on recognition.
Firstly, by replacing the bottom feature extraction structure in baseline with LSP, the algorithm recognition performance reaches 63.64\%.
The enhancement of the objective metrics demonstrates that the local scattered perceptron has meaningful perception of the target scattered information.
Based on this, the model recognition performance reaches 74.76\% and 77.78\% after adding SPR module and PKI module to the network.
Significant performance improvement over CNN-based backbone networks.
Moreover, the final complete method performance is improved by 18.59\% compared to the baseline AconvNet method.

\begin{table}[!t]
	\renewcommand\arraystretch{1.3}
	\caption{Module ablation experiments analysis}\label{tab5}
	\centering
	\setlength{\tabcolsep}{1.3mm}{\begin{tabular}{c c c c c}
			\hline
			LSP     & SPR  & PKI         &Overall accuracy &Performance Improvemetn\\ \hline
			$\times$    & $\times$    & $\times$   & 59.19\%      & -          \\
			$\times$    & $\checkmark$    & $\times$   & 72.73\%      & $\uparrow 3.35\%$          \\
			$\times$    & $\checkmark$  & $\checkmark$ & 75.69\%    &   $\uparrow 7.49\%$          \\
			\hline
			$\checkmark$    & $\times$  & $\times$  & 63.64\%      &  $\uparrow 1.75\%$            \\
			$\checkmark$   & $\checkmark$  & $\times$   & 74.76\%       & $\uparrow 3.35\%$          \\
			$\checkmark$   & $\checkmark$  & $\checkmark$  &77.78\%   & $\uparrow 9.36\%$       \\
			\hline
		\end{tabular}
	}
\end{table}

\subsubsection{Loss function validity analysis}

\begin{table*}[!t]
	\renewcommand\arraystretch{1.3}
	\caption{Loss function ablation experiments analysis}\label{tab6}
	\centering
	\setlength{\tabcolsep}{3.3mm}{\begin{tabular}{c c c c c c}
			\hline
			Method &w/o $L_{MPB}$     & w/o $L_{IPC}$  & w/o $L_{GDC}$         &w/o $L_{PSO}$  &Full model\\ \hline
			Accuracy   & 72.84\%    & 75.34\%   & 75.34\%      & 70.54\%    &77.78\%      \\
			
			\hline
		\end{tabular}
	}
\end{table*}
Then, the effect of different loss functions on the model is verified by an objective metric of recognition accuracy.
The quantitative results are shown in Table \ref{tab6}.
$\mathcal{L}_{\text {IPC}}$ is the crucial for local attention learning.
Otherwise, the powerful global sensory field of ViT drives itself to view the entire image, causing the model to degenerate into the color clustering model.

\begin{figure*}[!t]
	\centering \includegraphics[width=5.0in]{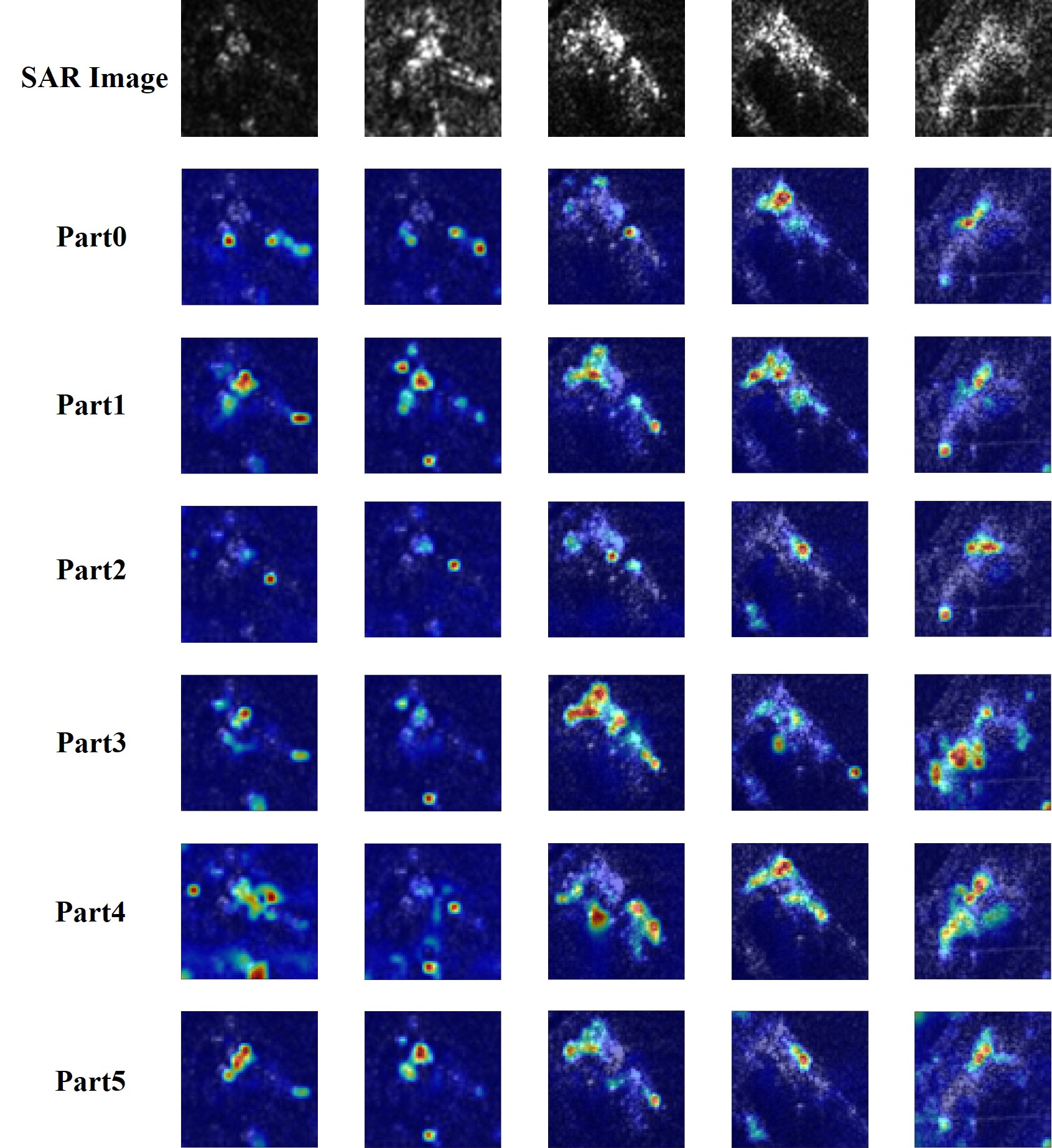}%
	\caption{Part feature visualization results on FAST-Vehicle.}
	\label{fig54}
\end{figure*}

\begin{figure*}[!t]
	\centering \includegraphics[width=5.0in]{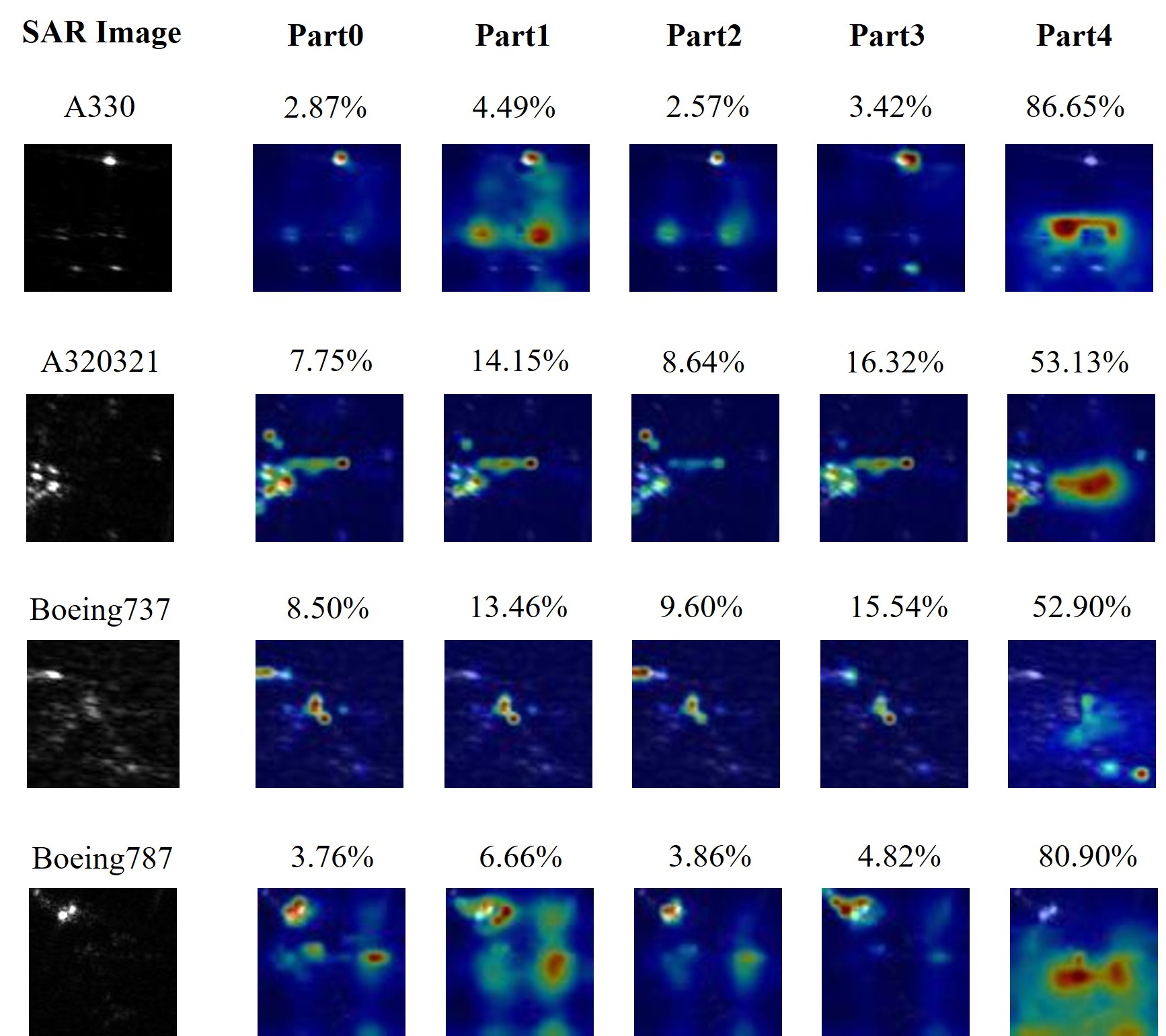}%
	\caption{Part feature visualization results on SAR-ACD.}
	\label{fig55}
\end{figure*}

Without $\mathcal{L}_{\text {MPB}}$, the model tends to prioritize $\mathcal{L}_{\text {IPC}}$ requirements.
As a result, the model converges to the mundane solution, recognizing almost all pixels as backgrounds.
$\mathcal{L}_{\text {GDC}}$ and $\mathcal{L}_{\text {PSO}}$ are also critical for target scattered part representation.
$\mathcal{L}_{\text {GDC}}$ ensures that the learned part representation explicitly contains the features of the corresponding part, rather than a series of meaningless vectors.
$\mathcal{L}_{\text {PSO}}$ expands the cosine distance between different partial representations, ensuring consistency in the discovery of the parts.

%
%
%
%

\subsection{Part visualization results}

Here, we visualize the results of the visualization of scattering parts on different datasets, as shown in Fig. \ref{fig54} and \ref{fig55}.
Specifically, the number of part divisions on the two datasets are set to 6 and 5.
Due to the diversity of SAR image variations, the proposed method is able to roughly extract information about different parts.

In particular, the contributions of different part features to the final results are exported on the SAR-ACD dataset.
Obviously, different parts contribute differently to the classification of different target categories.
Part4 has the largest percentage of all categories, which suggests that this part feature is the critical generic information of the aircraft and plays an undecisive role.
Ultimately, more robust and interpretable target recognition is achieved through the combined effect of information from multiple parts.

\subsection{Results of comparative experiments}
Subsequently, some representative methods are selected to perform experiments on FAST-Vehicle and SAR-ACD datasets to validate the performance of the proposed methods.

\begin{table*}[!t]
	\renewcommand\arraystretch{1.3}
	\caption{Algorithm performance under the FAST-Vehicle dataset\label{tabfv}}
	\centering
	\setlength{\tabcolsep}{2.3mm}
	\begin{tabular}{c c c c c c c} \hline
		&\multirow{2}{*}{Method}          &\multicolumn{5}{c}{Experiment condition}  \\
		\cline{3-7}
		& &EXP1 & EXP2 & EXP3 &EXP4 &EXP5 \\
		\hline
		\multirow{4}{*}{CNN-based} 		&AlexNet   & 52.95\%     & 59.73\%    & 45.66\%    & 64.32\%     & 53.79\%    \\
		
		&GooleNet     &  68.08\%     & 69.70\%    & 57.94\%   & 47.78\%     & 61.21\%    \\
		
		&VGGNet       & 69.76\%     & 70.49\%    & 54.80\%    & 47.15\%     & 56.78\%    \\
		
		&ResNet      &  67.15\%     & 70.91\%    & 55.60\%    & 67.56\%     & 63.55\%    \\
		
		\hline
		
		\multirow{6}{*}{SAR proprietary} &AconvNet     & 59.19\%      &  66.14\%     & 53.57\%     & 69.07\%   &62.16\%    \\
		
		&CBAM Net     &  67.30\%     & 68.05\%    &  48.58\%     & 53.27\%     & 49.76\%    \\
		
		&VIB Net     &  66.87\%     & 72.37\%    &  56.18\%      & 68.12\%      & 55.13\%     \\
		
		&FEC         &   60.12\%    &  67.97\%   &  53.72\%     &  63.12\%    &  54.15\%   \\
		
		&SDF-Net       &   62.23\%    &  64.58\%       &  53.26\%     &  68.47\%    &  59.73\%  \\
		
		&ASC-MACN       &  59.11\%   &  64.15\%   &  50.27\%     &   65.59\%   &  60.48\%    \\
		
		\hline
		&Proposed method        & 77.78\%     & 79.98\%    & 58.79\%     & 70.20\%     & 68.92\%    \\
		
		\hline
	\end{tabular}
\end{table*}
\subsubsection{Results on the FAST-Vehicle}
Specifically, the comparison methods consist of two main groups of generic vision methods and improved methods more suited to handle SAR ATR tasks.
Generic visual models are used to analyze the basic performance of the dataset.
While the improved methods are employed to compare the superiority of the proposed methods.
Specifically, we divide the FAST-Vehicle dataset meticulously to discuss the stability of the algorithm under various data distributions, as shown in Table. \ref{tabfv}.

EXP1, EXP2 and EXP3 data subsets, the only variable in the training and test sets is the degression.
In all experimental conditions, the proposed method achieved the highest or near-highest accuracy, demonstrating robust performance and strong generalization capabilities. 
Although traditional CNN architectures (e.g., AlexNet, GoogLeNet, VGGNet, ResNet) and certain SAR-specific networks (e.g., AconvNet, CBAM Net, VIB Net) attained relatively high accuracy in some experiments, they were generally outperformed by the proposed method.
The performance comparison with the CNN-based method leads to the following conclusions.
In EXP1 and EXP2, the proposed method (77.78\% and 79.98\%, respectively) achieved a 7\%–9\% performance improvement over the best-performing CNN networks (VGGNet or ResNet, approximately 70\%), clearly demonstrating its superiority. In EXP3, the proposed method reached 58.79\%, still surpassing GoogLeNet's 57.94\%, though the margin was relatively narrow, suggesting that this scenario is either more challenging or features a more demanding data distribution. In EXP4 and EXP5, the proposed method maintained the top performance (70.20\% and 68.92\%, respectively), showing an improvement of approximately 3–5 percentage points over the best-performing CNN, ResNet (67.56\% and 63.55\%).
In EXP1 and EXP2, compared to the highest accuracy achieved by the SAR-specific networks (generally ranging from 60\% to 72\%), the proposed method exhibits an improvement of 5–10 percentage points. In EXP3, although the absolute accuracy of the proposed method is not high (58.79\%), it still outperforms other SAR networks (e.g., VIB Net at 56.18\% and AconvNet at 53.57\%). In EXP4 and EXP5, the proposed method consistently ranks first (70.20\% and 68.92\%, respectively), demonstrating a clear advantage over the top-performing SAR networks (e.g., AconvNet, VIB Net, SDF-Net, which typically achieve around 65\%–69\%).

The proposed method achieved optimal or near-optimal results under all five experimental conditions, demonstrating strong feature learning and adaptability.
While traditional CNNs remain competitive with SAR-specific networks in some scenarios (e.g., EXP2 and EXP4), they generally underperformed compared to the proposed method, indicating that improvements tailored to SAR scenarios or specialized architectures yield significant performance gains.
The varying performance of different networks across experimental conditions is often linked to data distribution, scene characteristics, and the compatibility of model architectures. Further optimization could be pursued by leveraging domain-specific scene attributes.

\begin{table}[!t]
	\renewcommand\arraystretch{1.3}
	\caption{Algorithm performance on SAR-ACD dataset}\label{tabsa}
	\centering
	\setlength{\tabcolsep}{5.5mm}{\begin{tabular}{c c}
			\hline
			Method      & Accuracy   \\ \hline
			AlexNet   & 91.65\%    \\
			GoogleNet & 93.61\%  \\
			VGGNet    & 92.30\% \\
			ResNet    &92.79\% \\
			AconvNet  & 90.12\% \\
			Lm-BN-CNN & 87.34\% \\
			CBAM Net  & 92.66\% \\
			VIB Net   & 93.50\%  \\
			Proposed method & 95.43\%\\
			\hline
		\end{tabular}
	}
\end{table}

\subsubsection{Results on the SAR-ACD}

Table \ref{tabsa} shows the experimental results on the SAR-ACD dataset.
AlexNet, GoogleNet, VGGNet, and ResNet achieved accuracies of 91.65\%, 93.61\%, 92.30\%, and 92.79\%, respectively. 
Notably, GoogleNet attained the highest accuracy of 93.61\%, and the overall performance levels are relatively comparable, indicating that traditional CNN architectures can achieve robust performance on this dataset.
Among the networks specifically designed for SAR imagery, AconvNet and Lm-BN-CNN achieved accuracies of 90.12\% and 87.34\%, respectively, which are slightly inferior to the performance of traditional CNNs. In contrast, CBAM Net and VIB Net attained accuracies of 92.66\% and 93.50\%, respectively, which are comparable to or even slightly better than those of traditional CNN architectures, although the overall differences remain marginal.
The proposed method outperformed all others, achieving an accuracy of 95.43\%. Compared with the highest accuracy among traditional CNNs (GoogleNet at 93.61\%) and the top-performing SAR-specific network (VIB Net at 93.50\%), it demonstrated a significant improvement of approximately 1.8–1.9 percentage points, highlighting its superior feature extraction and discriminative capabilities.

Overall, the proposed method outperformed traditional CNNs and most SAR-specific networks on the SAR-ACD dataset, demonstrating its robustness and superiority in this domain.










\section{Conclusion}\label{}
We proposes a novel bottom-up scattering information perception network for more interpretable target recognition by constructing the proprietary interpretation network for SAR images.
Firstly, the localized scattering perceptron is proposed to replace the backbone feature extractor based on CNN networks to deeply mine the underlying scattering information of the target.
Then, an unsupervised scattering part feature extraction model is proposed to robustly characterize the target scattering part information and provide fine-grained target representation.
Finally, by aggregating the knowledge of target parts to form the complete target description, the interpretability and discriminative ability of the model is improved.
We demonstrate the state-of-the-art of the proposed method by validating on the vehicle and aircraft datasets FAST-Vehicle and SAR-ACD.

\printcredits

\bibliographystyle{cas-model2-names}

\bibliography{cas-refs}



\end{document}